\documentclass[runningheads]{llncs}

\usepackage[T1]{fontenc}
\usepackage{graphicx}
\usepackage{amsmath,amssymb}
\usepackage{booktabs}
\usepackage{url}
\usepackage{xcolor}
\usepackage{placeins}
\usepackage{bbding}

\newcommand{\Nones}{\ensuremath{N_{\mathrm{ones}}}}
\newcommand{\Bones}{\ensuremath{B_{\mathrm{ones}}}}
\newcommand{\Done}{\ensuremath{D_{\mathrm{ones}}}}
\newcommand{\Ozero}{\ensuremath{O[0]}}
\newcommand{\Oone}{\ensuremath{O[1]}}
\newcommand{\Rtwo}{\ensuremath{R^2}}

\begin{document}

\title{Represented Is Not Computed:\\A Causal Test of Candidate Algorithmic Intermediates in a Transformer}
\titlerunning{Represented Is Not Computed}

\author{Ishita Darade\inst{1}\orcidID{0009-0003-8489-563X}$^{*}$ \and Sushrut Thorat\textsuperscript{\Envelope}\inst{2,3}\orcidID{0000-0003-2276-5621}$^{*}$}
\authorrunning{I. Darade and S. Thorat}
\institute{MKSSS's Cummins College of Engineering for Women, Pune 411052, India\\
\email{ishitadarade1004@gmail.com}
\and
Institute of Cognitive Science, Osnabrück University, Osnabrück 49069, Germany\\
\and
Lossfunk, Indiranagar, Bengaluru, Karnataka 560038, India\\
\email{sushrut.thorat94@gmail.com}\quad $^{*}$Equal contribution. \Envelope\ Corresponding author.}

\maketitle

\begin{abstract}
Structured prompts require integrating components according to task-relevant relations. How a network implements this integration is often hard to judge in language or vision, where those relations are rarely specified precisely enough to define a candidate internal algorithm. Arithmetic offers a cleaner setting. We study a Transformer trained on base-digit extraction: given $N$, $B$, and $D$, it must report the coefficient of $B^D$ in the base-$B$ expansion of $N$. The closed-form solution, $\lfloor N/B^D \rfloor \bmod B$, provides explicit candidate algorithmic intermediates. Across three seeds, the model reaches 99.83\% exact-answer accuracy on held-out number--base intersections, establishing reliable task competence. Linear probes decode the intermediates, making staged arithmetic computation plausible. Causal tests then separate representation from use: within the localized route from the stream with $D$ as input to the output positions, behavior depends on early $D$-selective communication, independent of $N$ and $B$. Relatedly, a sparse circuit search finds mostly separate $N$, $B$, and $D$ routes that combine late rather than the staged route suggested by the probes. Thus, the model represents the intermediates that make the closed-form solution plausible, but the identified localized causal route does not transmit them to the output stream. This case shows that probe-based conclusions can diverge sharply from causal observations, even when explicit algorithmic hypotheses are available.
\end{abstract}

\section{Introduction}

To answer a structured prompt, a neural network must identify the relevant components and integrate them according to task-relevant relations. How it implements this integration is a central question in mechanistic interpretability. In natural language and vision, however, those relations are rarely specified precisely enough to define a candidate internal algorithm. This makes mechanistic interpretation difficult: before asking whether a model implements an algorithm, we need some account of the computation to be implemented~\cite{marr1982vision,marr1976understanding}.

Arithmetic gives a cleaner setting because the input-output function is known and candidate algorithms can be written down explicitly. Prior work has used arithmetic tasks to study mechanisms in Transformers and neural networks~\cite{nanda2023grokking,stolfo2023arithmetic,quirke2024arithmetic}. We extend this approach to a setting where the prompt does more than supply operands: it parameterizes the computation and specifies the requested readout. We study a Transformer trained on base-digit extraction, which requires combining a decimal number $N$, a base $B$, and a requested digit position $D$. The target is the coefficient of $B^D$ in the base-$B$ expansion of $N$:
\begin{equation}
    y = \left\lfloor \frac{N}{B^D} \right\rfloor \bmod B.
    \label{eq:target}
\end{equation}
For example, when $N=255$, $B=16$, and $D=0$, the answer is 15; when $D=1$, the answer is also 15; when $D=2$, the answer is 00. 

This expression suggests a staged algorithmic hypothesis: compute $B^D$, compute $N/B^D$, take the floor, and reduce modulo $B$. We use this staged decomposition as a candidate hypothesis about possible internal intermediates.

We train Transformers from scratch on this task and evaluate them autoregressively on held-out number--base intersections. We ask how the input components are used to form the answer. This setting lets us separate three questions that are often conflated: can the model solve the task, are quantities from the closed-form solution represented, and are those quantities the causal intermediates used to produce the answer? Decodability is often a starting point for mechanistic interpretation, but answering the last question requires causal tests.

We find that the model solves the task almost perfectly on held-out number--base intersections. Quantities aligned with the closed-form solution are linearly decodable from plausible streams and layers, making the staged algorithmic hypothesis tempting. But causal interventions reveal a different answer-forming circuit: when we trace how the requested digit position influences the output, the output streams read early $D$-selective information rather than the later quotient-like representations, and the broader circuit routes $N$, $B$, and $D$ mostly separately into the output streams before late integration.

The central contribution is therefore a dissociative observation: the model represents the quantities that make the algorithmic solution plausible, yet the identified \Done{}$\to O$ route does not carry them. Base-digit extraction gives us a setting where that divergence can be seen directly, turning an intuitive warning about probes into a directly testable dissociation\footnote{Training and analysis scripts are available at the project repository:\\
\url{https://github.com/ishita-darade10/represented-is-not-computed}}.

\begin{figure}[t]
    \centering
    \includegraphics[width=\linewidth]{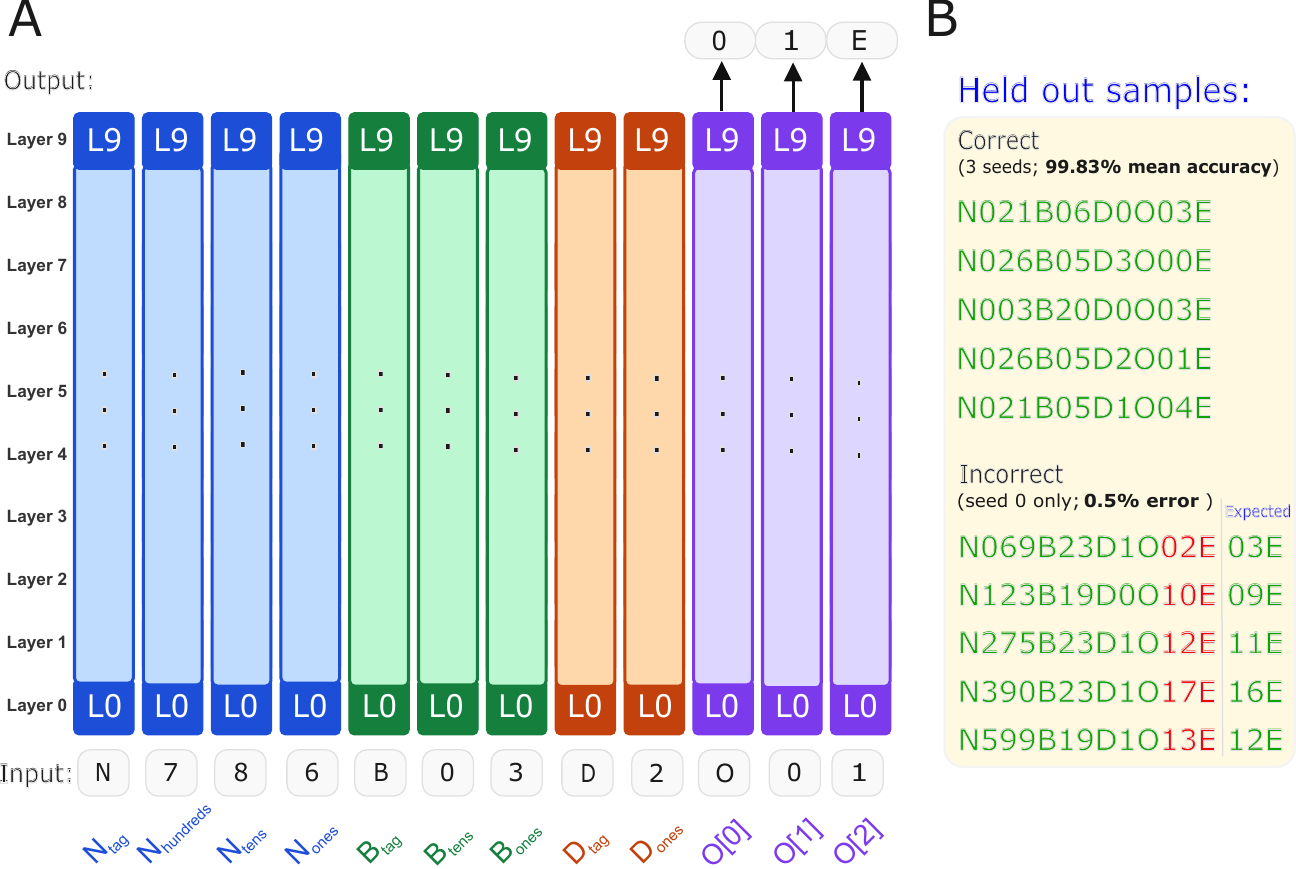}
    \caption{\textbf{Transformers solve base-digit extraction almost perfectly on held-out number--base intersections.} Across the three seeds, mean exact-answer accuracy was $99.83$\% ($95$\% CI: $99.50$--$100.00$\%). \textbf{A.} Canonical input order, named residual streams, and the three output streams used to generate the answer. \textbf{B.} Representative held-out examples answered correctly across the three seeds and representative errors from the network trained with seed $0$, the only network with non-perfect held-out accuracy; incorrect examples show the prediction followed by the expected answer.}
    \label{fig:task-performance}
\end{figure}

\section{Results}

The results follow the three questions above: task competence, representation of closed-form quantities, and causal use. The first two answers are yes; for the tested \Done{}$\to O$ route, the third is no. A stream is the sequence of residual states at a named token position across layers.

\subsection{Transformers learn the base-digit extraction task on held-out number--base intersections}

Before interpreting the model, we first asked whether it solved the task robustly enough to make mechanistic analysis meaningful. Each example contains $N$, $B$, and $D$; the target is the coefficient of $B^D$ in the base-$B$ expansion of $N$, written as a two-digit decimal answer (Section~\ref{sec:methods-task-definition}). We trained 10-layer decoder-only Transformers (three seeds) on $N\in\{0,\ldots,999\}$, $B\in\{2,\ldots,30\}$, and $D$ values from the ones digit through the highest base-$B$ digit position in the representation of $N$, plus one out-of-range query with target $00$ (Fig.~\ref{fig:task-performance}A). Training shuffled the tagged input segments ($N$xxx, $B$xx, $D$x) across all six permutations. Evaluation was autoregressive: the model received the prompt up to $O$ and greedily generated the two answer digits and end token; exact-answer accuracy uses the two answer digits.

The 10-layer models solved the held-out test sets ($N,B$ combinations; see Section~\ref{sec:methods-splits}) almost perfectly: across three independently trained 10-layer models, exact-answer accuracy on held-out number--base intersections was $99.83$\% ($95$\% CI: $99.50$--$100.00$\%; $99.50$\%, $100.00$\%, and $100.00$\% for seeds $0$, $42$, and $1337$, respectively; for examples, see Fig.~\ref{fig:task-performance}B). The corresponding token accuracies were also near ceiling: $99.98$\% ($95$\% CI: $99.95$--$100.00$\%) for the first answer digit and $99.83$\% ($95$\% CI: $99.50$--$100.00$\%) for the second answer digit. 

This establishes the behavioral target for interpretation. The model learned a reliable mapping across held-out $N,B$ combinations within the trained numerical range; it was not merely memorizing the evaluated number--base pairs as training examples, although this does not prove unbounded algorithmic generalization.

\enlargethispage{\baselineskip}
\subsection{Closed-form quantities are linearly decodable from the residual streams}

Having established that the models solve the task, we next asked whether the closed-form solution leaves a representational trace inside the network. The formula suggests a staged algorithm: compute $B^D$, compute $N/B^D$, take the floor, and then reduce modulo $B$. If the Transformer uses something like this algorithm, these quantities should be accessible from the residual stream, perhaps at staged depths. We therefore trained linear probes: ordinary linear readouts on frozen activations from pooled held-out validation and test examples (Section~\ref{sec:methods-linear-probing}). The targets were $B^D$, $N/B^D$, $\lfloor N/B^D\rfloor$, and $\lfloor N/B^D\rfloor \bmod B$. As in the probing literature, linear probe success means accessibility to a simple readout~\cite{alain2016understanding,hewitt2019probes,belinkov2022probing,elazar2021amnesic}.

\begin{figure}[t]
    \centering
    \includegraphics[width=0.97\linewidth]{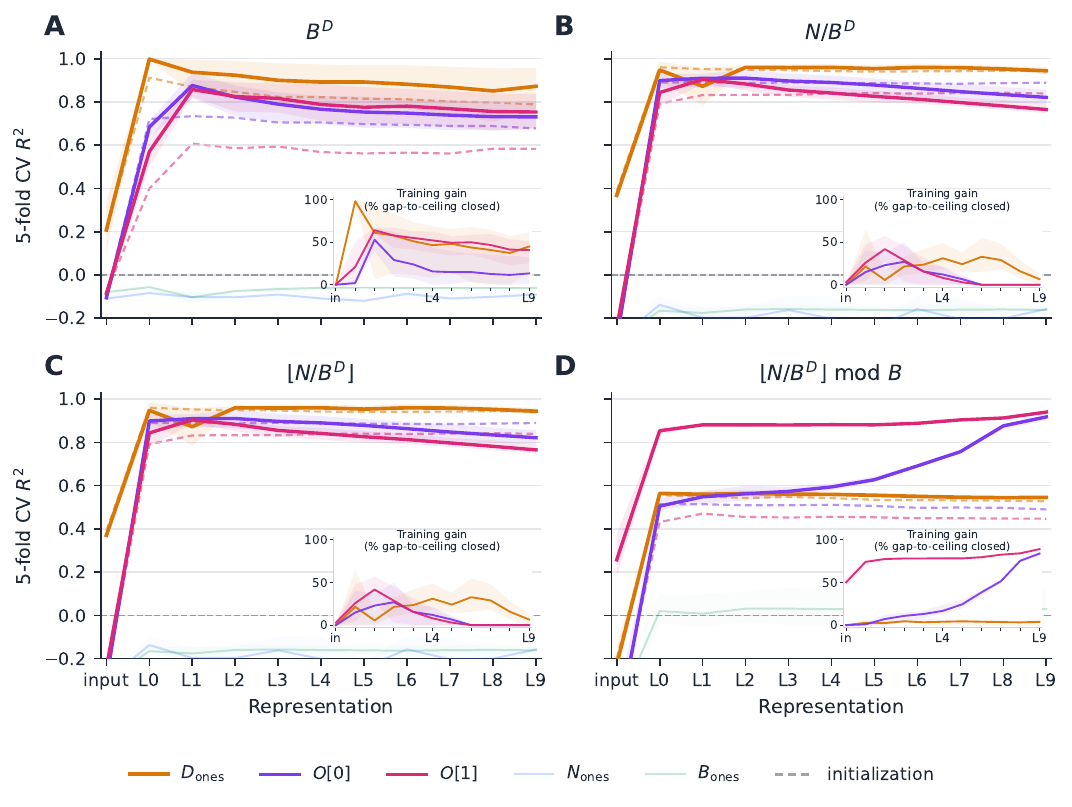}
    \caption{\textbf{Closed-form quantities are decodable, making a staged algorithmic hypothesis representationally plausible.} Solid lines show trained-model mean 5-fold cross-validated \Rtwo{} with shaded $95$\% confidence intervals across seeds; dashed lines show matched initialization baselines for the same streams. Insets show the training-added gain, measured as the percent of the initialization-to-ceiling \Rtwo{} gap closed by training. The \Done{} stream is strongest for $B^D$, $N/B^D$, and $\lfloor N/B^D\rfloor$, with substantial decodability already present at initialization; the final answer quantity, $\lfloor N/B^D\rfloor \bmod B$, decodes strongest late in the output streams. \Nones{} and \Bones{} are controls; negative \Rtwo{} means worse than predicting the held-out fold mean.}
    \label{fig:probes}
\end{figure}

The closed-form quantities were strongly decodable (Fig.~\ref{fig:probes}). $B^D$ was best decoded from the \Done{} stream at layer $0$, with mean cross-validated \Rtwo{} $= 1.00$ ($95$\% CI: $1.00$--$1.00$). The quotient-like quantities were also strongest in the \Done{} stream: both $N/B^D$ and $\lfloor N/B^D\rfloor$ reached mean \Rtwo{} $= 0.96$ ($95$\% CI: $0.95$--$0.97$) at layer $2$, though they were already highly decodable at earlier and later layers. The final answer quantity, $\lfloor N/B^D\rfloor \bmod B$, was strongest late in the output stream, with mean \Rtwo{} $= 0.94$ ($95$\% CI: $0.93$--$0.95$) from \Oone{} at layer $9$. Together, these results make the staged algorithmic hypothesis representationally plausible: the quantities suggested by the closed-form solution are linearly accessible in streams and layers broadly consistent with the proposed staged algorithm.

A possible concern is that these targets might be recoverable from the input scalars, with a linear model, due to correlations. To check this, we regressed each target directly on raw $N$, $B$, and $D$ over the same pooled held-out validation and test examples. This reached only mean \Rtwo{} $= 0.25$ ($95$\% CI: $0.16$--$0.35$) for $B^D$, $0.50$ ($95$\% CI: $0.39$--$0.56$) for both quotient-like targets, and $0.33$ ($95$\% CI: $0.28$--$0.37$) for the answer. Thus, raw scalar correlations explain only part of the linear predictability; the residual streams make the closed-form quantities substantially more accessible.

Another concern is that this accessibility might already be present before training. To test this, we reconstructed initialized networks with the same seeds, architectures, and held-out splits, and repeated the probing analysis. The intermediate targets were substantially decodable even at initialization, in some cases nearly as well as after training (dashed lines in Fig.~\ref{fig:probes}). Training nevertheless changed this structure systematically (insets in Fig.~\ref{fig:probes}). For $B^D$ in \Done{} at layer $0$, training nearly closed the initialization-to-ceiling \Rtwo{} gap ($98.00$\%; $95$\% CI: $96.00$--$99.20$\%). For quotient-like quantities, the training-added gain in \Done{} peaked later, at layer $6$ ($32.80$\%; $95$\% CI: $15.30$--$55.00$\%). The answer quantity decoding gained most in the output streams, peaking late at \Oone{} layer $9$ ($89.10$\%; $95$\% CI: $87.20$--$90.50$\%). Thus, the probes reveal both pre-training accessibility and training-shaped structure: early \Done{} access to $B^D$, later refinement of quotient-like quantities, and late output-side answer formation.

\enlargethispage{\baselineskip}
\subsection{The output relies especially on early \Done{} communication, not primarily on later quotient-like states}

The probing results suggest that the first three closed-form quantities are especially accessible from the \Done{} stream. We next asked whether the output streams use \Done{} in a way that follows this probe profile. If the output streams mainly read richer quotient-like information from \Done{}, then causal sensitivity should not be confined to the earliest \Done{} readouts. If instead the behaviorally effective contribution of \Done{} is closer to digit-position information, performance should be especially sensitive to early \Done{} communication.

We therefore performed targeted attention ablations from the output streams to \Done{}, blocking only the \Ozero{} and \Oone{} attentional reads from \Done{} at selected layers while leaving all other attention reads and residual streams unchanged (Section~\ref{sec:methods-attention-ablation}). We measured exact-answer accuracy in cumulative shallow-to-deep and deep-to-shallow sweeps, which distinguish early from late dependence.

\begin{figure}[t]
    \centering
    \includegraphics[width=\linewidth]{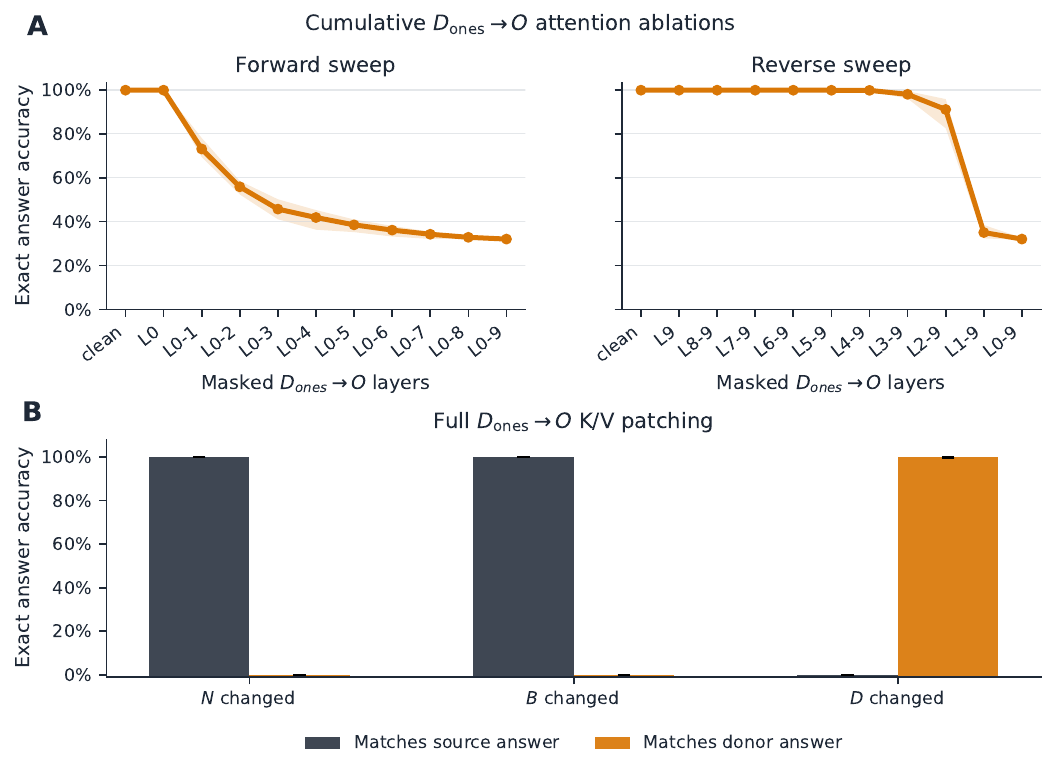}
    \caption{\textbf{The behaviorally effective information carried by the \Done{}$\to O$ route is early and $D$-selective.} \textbf{A.} Cumulative ablations show that output behavior is most sensitive once attention in the early \Done{}$\to O$ ($O[0]$ and $O[1]$) route is removed. \textbf{B.} Full-route donor key/value patching tests what that route carries: substituting donor \Done{} key/value vectors transfers behavior only when the donor differs in $D$, not when it differs in $N$ or $B$; dark bars denote source-answer matches and orange bars donor-answer matches. Shading and error bars show $95$\% confidence intervals across seeds.
    }
    \label{fig:causal-tests}
\end{figure}

The output streams depended most strongly on early \Done{} communication (Fig.~\ref{fig:causal-tests}A). In the forward sweep, masking layer $0$ alone had no effect: mean accuracy remained $99.83$\% ($95$\% CI: $99.50$--$100.00$\%). But masking layers $0$--$1$ caused a large drop to $73.08$\% ($95$\% CI: $69.36$--$78.08$\%), and masking layers $0$--$2$ reduced performance further to $55.85$\% ($95$\% CI: $52.36$--$58.50$\%). Masking all \Done{}-to-output attention across layers reduced performance to $32.06$\% ($95$\% CI: $31.72$--$32.62$\%).

The reverse sweep gave the complementary pattern. Masking the deepest layers barely affected performance. Accuracy remained $99.75$\% ($95$\% CI: $99.45$--$100.00$\%) even when layers $4$--$9$ were masked. Performance only began to drop substantially when the sweep reached earlier layers: masking layers $2$--$9$ reduced accuracy to $91.04$\% ($95$\% CI: $82.50$--$95.79$\%),\footnote{Seed $0/42/1337$: $95.79/94.82/82.50$\%.} and masking layers $1$--$9$ reduced it to $35.06$\% ($95$\% CI: $32.36$--$38.47$\%), close to the full-ablation level.

This pattern is hard to reconcile with a simple staged closed-form computation in which the output streams mainly read quotient-like intermediates from \Done{} states. The quotient-like quantities are already decodable from early \Done{} states and become strongest around layer \(2\), but the causal sensitivity does not follow the probe profile in a simple way. In the reverse sweep, performance stays near ceiling until layer \(1\) is included, then drops close to the full-ablation level. Because layer \(1\) attention reads the layer \(0\) \Done{} state, the most sensitive readout is tied to the earliest \Done{} communication rather than selectively to the later layers where the richer closed-form quantities are maximally decodable. Later \Done{} states still contribute, as the forward sweep continues to fall when additional layers are removed. The conclusion is therefore narrower: decodability marks informative states along the route, but it does not by itself identify which information the output streams use causally.

\enlargethispage{\baselineskip}
\subsection{Layer-1 \Done{}-to-output patching steers the answer toward the donor}

Early \Done{}-to-output communication is the most sensitive part of the route, but ablations remove a route and can put the model in an unnatural state. We therefore used patching, which leaves the computation otherwise intact while replacing what one route provides with donor information~\cite{meng2022rome,goldowskydill2023path,zhang2024patching}. If the patched output shifts toward the donor answer, that route has causal control during otherwise normal computation.

We constructed source--donor pairs matched on $N$ and $B$ but differing in $D$, and patched donor key/value vectors from the \Done{} token position into the output-query rows while leaving the source \Done{} residual stream unchanged. We patched only layer $1$, because the ablation analysis identified it as the most sensitive point. This asks whether the output streams behave as if they read the source $D$ or the donor $D$ (Section~\ref{sec:methods-kv-patching}).

Patching only the layer $1$ route shifted behavior toward the donor answer. Across the three $10$-layer seeds, source-exact accuracy fell to $25.11$\% ($95$\% CI: $13.58$--$45.18$\%), while donor-exact accuracy rose to $67.63$\% ($95$\% CI: $42.72$--$81.13$\%).\footnote{Seed $0/42/1337$: source-exact $13.58/16.59/45.18$\%; donor-exact $81.13/79.03/42.72$\%.}

The transfer is substantial but incomplete, showing that $D$-dependent control is distributed across the route rather than isolated to one layer. Still, this result confirms that the early route identified by ablation has substantial causal control during normal operation.

\subsection{The full \Done{}-to-output route carries behaviorally effective information about $D$, not $N$ or $B$}

The previous analyses show that \Done{}-to-output communication matters, especially early. But they do not yet tell us what behaviorally effective information the output positions receive from that attention route. This distinction is crucial. The probes show that \Done{} contains quantities involving all three factors, $N$, $B$, and $D$. If the output streams read a closed-form intermediate such as $N/B^D$ or $\lfloor N/B^D\rfloor$ from this route, then patching donor information along this route should transfer behavior when the donor differs in $N$, $B$, or $D$. A $D$-only-routing account makes a different prediction: only donor changes in $D$ should transfer.

We tested this directly. For each clean-correct held-out source example, we selected three donors: one differing only in $N$, one only in $B$, and one only in $D$. We then patched the donor \Done{} key/value readout into the output query rows at every layer while leaving the source \Done{} residual stream unchanged. The same retained sources were used across the three donor conditions, so condition differences cannot be explained by different source populations (Section~\ref{sec:methods-kv-patching}).

The patching result was selective (Fig.~\ref{fig:causal-tests}B). When the donor differed only in $N$, full-route patching had no effect: the models remained $100.00$\% source-exact ($95$\% CI: $100.00$--$100.00$\%). When the donor differed only in $B$, the same was true: $100.00$\% source-exact ($95$\% CI: $100.00$--$100.00$\%). But when the donor differed only in $D$, behavior flipped almost perfectly: source-exact accuracy fell to $0.00$\% ($95$\% CI: $0.00$--$0.00$\%), and donor-exact accuracy rose to $99.84$\% ($95$\% CI: $99.53$--$100.00$\%). Thus, the behaviorally effective information carried from \Done{} to the output streams is highly selective for $D$.

This gives the key dissociation: the \Done{} residual stream makes $N$- and $B$-dependent closed-form quantities linearly available, but the behaviorally effective information read from \Done{} by the output streams is $D$-dependent. The represented quantities therefore do not describe the information carried along the \Done{}$\to O$ route.

\begin{figure}[t]
    \centering
    \includegraphics[width=\linewidth]{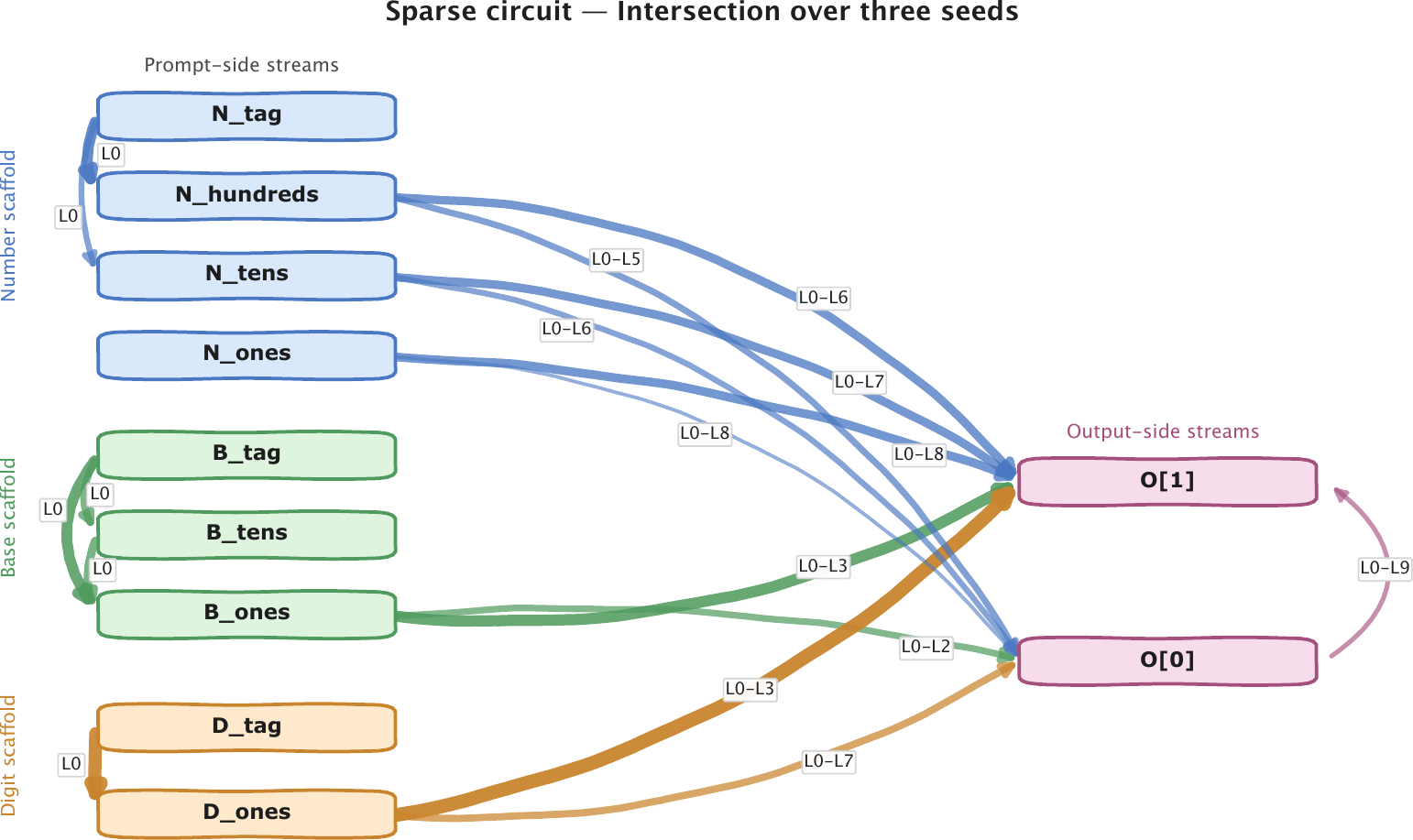}
    \caption{\textbf{Sparse circuit search finds factorized routes that converge at the output.} The across-seed intersection circuit contains separate local scaffolds for $N$, $B$, and $D$, each routing into the output streams. The retained \Ozero{}$\to$\Oone{} edge indicates that the two output positions are coupled. Edge labels give retained layer ranges. Across seeds, kept-only circuits achieved $93.40$\% held-out exact-answer accuracy ($95$\% CI: $90.83$--$97.68$\%) while retaining $28.33$\% of candidate source-target stream relations ($95$\% CI: $27.85$--$28.57$\%) and $20.25$\% of their candidate layer-edges ($95$\% CI: $19.47$--$21.20$\%); $17$ of $18$ retained relations ($94.44$\%) were shared across all three seeds.}
    \label{fig:sparse-circuit}
\end{figure}

\subsection{A sparse circuit search reveals mostly factorized routing into the output streams}

The \Done{} patching results rule out one natural route for staged closed-form computation: \Done{} does not send a behaviorally effective $N$- and $B$-dependent intermediate to the output streams. A broader possibility remains: $N$, $B$, and $D$ might interact elsewhere before output-side readout. To test this, we searched for a sparse circuit: a subset of attention routes sufficient for most task performance.

We used a greedy right-to-left search in the style of prior Transformer-circuit work~\cite{elhage2021framework,wang2023interpretability,conmy2023automated}. Starting from the output streams, it retained source-to-destination attention routes whose removal damaged validation accuracy; each seed's kept-only circuit, with non-retained incoming routes masked, was then tested autoregressively on that seed's test split (Section~\ref{sec:methods-sparse-circuit}). This is a sufficient-circuit search, not a proof of uniqueness or minimality.

The discovered circuits preserved most of the model's performance while retaining a small fraction of candidate routes. Across the three $10$-layer seeds, the held-out exact accuracy for the full network was $99.83$\% ($95$\% CI: $99.50$--$100.00$\%), and the kept-only circuit achieved $93.40$\% exact accuracy ($95$\% CI: $90.83$--$97.68$\%). The circuits retained on average $28.33$\% of possible source-target relations ($95$\% CI: $27.85$--$28.57$\%) and $20.25$\% of candidate layer-edges ($95$\% CI: $19.47$--$21.20$\%).

The route-level structure was highly stable across independently trained models: the across-seed intersection contained $17$ of the $18$ relations in the union, or $94.44$\% intersection-over-union. We summarize this intersection circuit in Fig.~\ref{fig:sparse-circuit}. Seeds $42$ and $1337$ had identical relation sets, and seed $0$ differed only by the absence of the $B_{\mathrm{tens}}\to$\Ozero{} relation. This shared circuit was mostly factorized before the output streams: local $N$ routes connected the number tag to number digit streams and routed the number digits to both outputs; local $B$ routes connected the base tag to base digit streams, included a $B_{\mathrm{tens}}\to$\Bones{} scaffold, and routed \Bones{} to both outputs; and the $D$ route connected $D_{\mathrm{tag}}\to$\Done{} and then \Done{}$\to$\Ozero{} and \Done{}$\to$\Oone{}. The retained \Ozero{}$\to$\Oone{} route shows that the output streams are not independent.

We tested threshold dependence with a $3\times3$ sweep across seeds ($27$ checks total; Section~\ref{sec:methods-sparse-circuit}). Relation sets averaged $92.77$\% overlap with the Fig.~\ref{fig:sparse-circuit} shared circuit, and $21/27$ checks contained all $17$ shared relations. The sweep identified no stable prompt-side relation combining the $N$, $B$, and $D$ scaffolds before output-side readout; only $N_{\mathrm{tag}}\to B_{\mathrm{tag}}$ appeared at the loosest threshold for seed $1337$ ($3/27$ checks). Together, these checks suggest that the factorized routing pattern is not an idiosyncratic artifact of one seed or one threshold setting.

Thus, the broader circuit supports the late-integration picture. We find little evidence that the main $N$, $B$, and $D$ pathways are first combined into a single closed-form intermediate in the prompt-side streams. Instead, the model routes the factors through mostly separate scaffolds into the output side, where the final answer is formed by output-side computation not fully decomposed here.

\raggedbottom
\section{Discussion}

We asked how a Transformer forms an answer in base-digit extraction, where the prompt parameterizes the computation and requested readout. The closed-form solution gives a natural staged hypothesis: compute $B^D$, $N/B^D$, and $\lfloor N/B^D \rfloor$, then produce the requested digit. This looked plausible representationally: the quantities were linearly decodable from plausible streams and layers. Causally, however, early \Done{}$\to O$ communication was especially influential, full-route patching transferred behavior only when $D$ changed, and the sparse circuit routed $N$, $B$, and $D$ mostly separately into output streams. Thus, \Done{} makes candidate algorithmic quantities decodable without passing them to the output streams as the main causal intermediates.

The claim is deliberately narrower than saying that the model never implements the closed-form computation. We do not decompose the output streams, and we do not rule out nonlinear integration there. What we test is whether the specific quantities that motivate the staged algorithmic hypothesis are the main causal intermediates carried by the routes we identify. They are not. Here, ``computed'' means causally used in forming the answer, not merely recoverable from an activation. The key dissociation is not that a digit-position stream carries $D$-dependent information, but that richer $N$- and $B$-dependent quantities are decodable from \Done{} yet do not describe what the output reads from \Done{}. In causal-abstraction terms, the closed-form solution is useful as a representational hypothesis, but the localized \Done{}$\to O$ route does not faithfully implement the hypothesized intermediates~\cite{geiger2021causal,geiger2022inducing,geiger2024alignments}. In this sense, ``represented is not computed''.

The initialization probes sharpen this point. Some nontrivial closed-form quantities were already highly decodable before training, even though a linear model on raw $N$, $B$, and $D$ explained much less variance. Training then reshaped this accessibility in task-relevant ways: early \Done{} access to $B^D$, later quotient-like refinement, and late output-side answer decoding. Thus, high decodability need not mark a learned causal intermediate~\cite{hewitt2019probes,belinkov2022probing,elazar2021amnesic}. It can reflect accessibility supplied by architecture, tokenization, and dataset geometry, later sculpted by training. Because Transformer blocks and feedforward weights are shared across token positions, transformations supporting output-side integration may make related quantities represented elsewhere without routing them forward for behavior. This possibility is also compatible with the broader view that neural representations can make many features linearly recoverable without assigning each a separate modular computational role~\cite{elhage2022toy}.


The remaining open piece is the output-side computation itself. We localize what reaches the output streams: $N$, $B$, and $D$ arrive largely through separate routes, and \Done{}$\to O$ supplies behaviorally effective $D$-selective information rather than a quotient-like intermediate. But we do not yet decompose how $O[0]$ and $O[1]$ combine these inputs to produce the answer. The final computation may still be structured and arithmetic; it is just not organized as the staged prompt-side transmission suggested by the probes.

This work complements work on Transformer circuits~\cite{elhage2021framework,olsson2022induction,wang2023interpretability}, route-specific patching~\cite{goldowskydill2023path,zhang2024patching}, and arithmetic mechanisms~\cite{stolfo2023arithmetic,quirke2024arithmetic,nikankin2024arithmetic}. Our setting is unusually explicit: the target function is known, the candidate intermediates are fixed by the closed-form solution, the model is trained from scratch, and held-out performance is near ceiling. The mismatch is therefore not a failure of task competence or a vague target algorithm. The candidate intermediates are clear, linearly available, and still not the causal intermediates carried by the tested routes. This connects to evidence that language models can solve arithmetic using heuristic rather than clean algorithmic routes~\cite{nikankin2024arithmetic}; here, the warning appears in a bounded setting where the candidate algorithm can be tested directly.

The practical lesson is not to abandon probing. Probes did useful work here: they identified a plausible algorithmic hypothesis. The next step was to ask whether downstream routes actually used the probed quantities~\cite{alain2016understanding,hewitt2019probes,belinkov2022probing,vig2020causal,geiger2021causal,wang2023interpretability,conmy2023automated}. The same distinction is familiar in cognitive neuroscience: decoding and representational similarity show what information is present, whereas perturbations test whether that information is used for behavior~\cite{norman2006mvpa,kriegeskorte2008rsa,weichwald2015causal,jazayeri2017neural}.

This architectural possibility also motivates comparisons with recurrent predictive models, which may place stronger pressure on maintaining task-relevant state across time~\cite{ventura2026minimal}.

In sum, even when a candidate algorithm is explicit and its quantities are decodable, mechanistic explanation requires showing how those quantities are used causally.

\enlargethispage{2\baselineskip}
\flushbottom
\section{Materials and Methods}

\subsection{Task definition}
\label{sec:methods-task-definition}

The task was designed so each prompt specifies both operands and the requested readout. Each example presents a decimal integer $N$, a base $B$, and a digit position $D$. The target is the coefficient of $B^D$ in the base-$B$ representation of $N$ (Eq.~\ref{eq:target}), emitted as a two-digit decimal string followed by an end token:
\begin{verbatim}
N 255 B 16 D 0 O 15 E
N 255 B 16 D 1 O 15 E
N 255 B 16 D 2 O 00 E
\end{verbatim}
The domain was $N \in \{0,\ldots,999\}$ and $B \in \{2,\ldots,30\}$. For each $(N,B)$, $D$ indexes the requested base-$B$ digit, with $D=0$ the ones digit. Queries included every valid position and the next out-of-range query (target $00$), except that $D=10$ was omitted rather than clipped for $B=2$, $N\geq 512$. For $N=0$, $D=0$ remained in range; capping $D\leq 9$ kept $D$ a one-character token. Base digits were emitted as two decimal characters, so values above $9$ were written as $10$, $11$, and so on, not hexadecimal-style letters.

\subsection{Splits and analysis datasets}
\label{sec:methods-splits}

To test held-out number--base intersections rather than repeated evaluated pairs, the main networks used a \texttt{by\_NB\_intersection} split. Validation examples used validation $N$ values paired with validation $B$ values; test examples used test $N$ values paired with test $B$ values; all remaining examples were assigned to training. Because held-out identities and realized sample counts were seed-specific, every analysis reconstructed them from the \texttt{split\_info} object stored with that seed rather than regenerating them from configuration. The policy selected up to $10$\% of $N$ values each for validation and test and up to $20$\% of $B$ values each for validation and test, realizing $100/1000$ $N$ values and $5/29$ $B$ values in each held-out split.

Held-out test examples were used for task evaluation, attention ablation, both key/value patching analyses, and final sparse-circuit evaluation. Linear probes used pooled held-out validation and test examples, without training data. Sparse-circuit discovery used each seed's held-out validation split, after which the circuit was frozen before test evaluation. Unless stated otherwise, evaluation examples used the canonical $N\ldots B\ldots D\ldots O$ order.

\subsection{Token streams nomenclature}

We refer to residual streams by token role: $N_{\mathrm{tag}}$, $N_{\mathrm{hundreds}}$, $N_{\mathrm{tens}}$, and \Nones{} for the number field; $B_{\mathrm{tag}}$, $B_{\mathrm{tens}}$, and \Bones{} for the base field; and $D_{\mathrm{tag}}$ and \Done{} for the digit-position field. For the output field, \Ozero{} is the stream at the output marker used to predict the first answer digit, and \Oone{} is the stream at the first generated answer digit used to predict the second.

\subsection{Model architecture and training}
\label{sec:methods-arch-training}

To keep the mechanism interpretable while preserving standard attention dynamics, the main model was a GPT-style decoder-only Transformer~\cite{vaswani2017attention} with $10$ layers, width $384$, $12$ attention heads, feed-forward width $1536$, dropout $0.01$, learned token and positional embeddings, and maximum sequence length $13$. Models were trained for $1000$ epochs with AdamW (learning rate $5\times10^{-4}$, weight decay $0.01$, $\beta=(0.9,0.95)$), linear warmup over the first $5$\% of steps, and batch size $2048$. Training expanded each training example across all six field-order permutations, forcing use of field tags rather than fixed absolute positions. The main condition used three independently trained $10$-layer networks (seeds $0$, $42$, and $1337$), using each seed's best validation epoch. A $5$-layer companion under the same permutation condition (seed $1337$) showed the same qualitative pattern: $99.89$\% held-out exact accuracy; $64.46$\% after masking \Done{}$\to O$ layers $0$--$1$; $21.42/63.43$\% source/donor exact after layer-$1$ patching; and full-route selectivity for $D$ ($N$: $100.00/0.00$\%, $B$: $100.00/0.00$\%, $D$: $0.00/99.87$\%).

\subsection{Autoregressive evaluation and statistical summaries}
\label{sec:methods-evaluation-stats}

Because the answer has two digits, all behavioral analyses used greedy autoregressive decoding. The model received the prompt through $O$, generated the first answer digit, then conditioned on that digit to produce the second answer digit and end token. We report exact-answer accuracy when both answer digits match their targets, and per-digit token accuracy when needed; the end token is generated but excluded from answer metrics.

Whenever the paper reports a mean across the three main seeds, the seed is the replicate unit. We report arithmetic means and two-sided percentile-bootstrap $95$\% confidence intervals over seed-level means, using $100{,}000$ resamples and bootstrap RNG seed $20260517$. With only three networks, these intervals are descriptive summaries of across-seed variation rather than finely estimated population intervals.

\subsection{Linear probing}
\label{sec:methods-linear-probing}

To measure linear accessibility, we collected residual-stream activations autoregressively from pooled held-out validation and test examples for each seed. Each model ran through $O$, generated the first answer digit greedily, then conditioned on that digit before collecting the stream used to predict the second digit. We probed $B^D$, $N/B^D$, $\lfloor N/B^D\rfloor$, and $\lfloor N/B^D\rfloor\bmod B$ for each stream and layer. We analyzed the input residual and every post-block residual for all named streams except $O[2]$, which only predicts the end token. Probes were ordinary least-squares regressions with intercepts; features were standardized on training folds only, and performance was shuffled $5$-fold cross-validated \Rtwo{} within the pooled held-out set, with fold seed $20260517$.

Initialization baselines repeated the protocol for each seed's architecture and held-out split with weights at initialization. Fig.~\ref{fig:probes} insets report \(\Delta^+ = 100\times\max(0,(R^2_t-R^2_i)/(1-R^2_i))\); before clipping, mean $R^2_t-R^2_i$ across the shown combinations ranged from $-0.08$ to $0.75$.

\subsection{Attention ablation}
\label{sec:methods-attention-ablation}

The ablation analysis asks where the output streams need access to \Done{}. We masked attention from output queries to the \Done{} token at selected layers: \Ozero{} during first-digit generation and \Oone{} during second-digit generation. Selected pre-softmax logits were set to $-10^9$, yielding zero post-softmax attention after renormalization; all other edges and residual streams were unchanged. This route-specific intervention follows mediation and path-patching work~\cite{vig2020causal,goldowskydill2023path,zhang2024patching}. We ran cumulative shallow-to-deep ($0$, $0$--$1$, $0$--$2$, \ldots) and deep-to-shallow ($9$, $8$--$9$, $7$--$9$, \ldots) sweeps on each seed's full held-out test split.

\subsection{Key/value route patching}
\label{sec:methods-kv-patching}

Patching tests what a route carries without deleting it. The source is the prompt being evaluated; the donor supplies replacement key/value vectors; source-match or donor-match indicates which answer the patched output equals. Both examples were first run normally. The patched run used the source prompt, kept output queries source-like, and substituted donor-derived \Done{} key/value vectors only in the output-query rows (\Ozero{} for the first digit, \Oone{} for the second). At layer $L$, patched key/value vectors came from the residual entering layer $L$; all non-output query rows, non-\Done{} token slots, and the source \Done{} trajectory stayed source-like. This isolates the intended \Done{}$\to O$ readout.

For layer-$1$ patching, we used all valid ordered held-out test source--donor pairs that shared $(N,B)$, differed in $D$, and had different two-digit answers; donor \Done{} key/value vectors were substituted at layer $1$ only. For full-route patching, each clean-correct held-out source was paired within the same test split with one donor differing only in $N$, one only in $B$, and one only in $D$; each donor had a different answer. One donor per condition was selected deterministically with RNG seed $20260517$, so all conditions used the same source set. We report whether patched outputs match the source or donor answer.

\subsection{Sparse circuit search}
\label{sec:methods-sparse-circuit}

To ask how task information reaches the output, we used a greedy right-to-left attention-route search~\cite{elhage2021framework,wang2023interpretability,conmy2023automated}. Discovery used each seed's full held-out validation split; the frozen circuit was evaluated autoregressively on that seed's test split. Starting with \Ozero{} and \Oone{} as destinations, we masked each candidate source--destination attention relation over layer prefixes and measured the validation exact-answer drop. Scoring was fully autoregressive: the first digit was generated under intervention and then used to generate the second, so retention reflected full two-digit behavior rather than a teacher-forced \Oone{} diagnostic. If a source stream was retained, it became a new destination, growing the circuit backward from the output.

For \Ozero{}, candidate sources were prompt streams; for \Oone{}, they were prompt streams plus \Ozero{}, so \Ozero{}$\to$\Oone{} was discovered rather than added by hand. For prompt-side destinations, candidate sources were earlier prompt-side streams only. For each candidate relation $s\to d$, we tested cumulative layer prefixes from layer $0$ to the deepest layer at which $d$ was retained downstream. For each prefix $0$ through $k$, only the attention edge $s\to d$ was masked, by setting pre-softmax logits to $-10^9$. A relation is a source--destination pair ignoring layer; a layer-edge is that relation at one layer.

A relation was retained when its validation ablation curve met the greedy rule: the first accuracy drop exceeded $0.02$, later extensions caused additional drops exceeding $20$\% of that first drop, and the scan stopped after two consecutive extensions failed this criterion. The retained prefix $0$ through $k^*$ defined the layers kept for that relation. The threshold sweep crossed first-drop thresholds $\{0.01,0.02,0.05\}$ with later-drop fractions $\{0.10,0.20,0.30\}$ and the same stopping rule. When an upstream relation was retained, we recorded the downstream stream through which it affected performance.

After discovery, retained prefixes were unioned into one seed-specific circuit. In kept-only evaluation, all non-retained admissible incoming routes into constrained destinations were masked at the corresponding layers, and the sparse circuit was evaluated autoregressively on that seed's held-out test split. We report clean and kept-only exact-answer accuracy, retained relation fraction, retained layer-edge fraction, and relation-level overlap. Intersection-over-union is set overlap over retained source-target relations, $|\cap_i R_i|/|\cup_i R_i|$, across seeds; threshold-sweep overlap compares each threshold-specific set with the $17$-relation Fig.~\ref{fig:sparse-circuit} shared set. The search identifies a sparse sufficient circuit, not a formally minimal or unique one.

\enlargethispage{2\baselineskip}
\begin{credits}
\subsubsection{\discintname} The authors have no competing interests to declare that are relevant to the content of this article.
\end{credits}

\end{document}